# Artificial intelligence application in lymphoma diagnosis: from Convolutional Neural Network to Vision Transformer

**Daniel Rivera, Jacob Huddin, Alexander Banerjee, Rongzhen Zhang, Brenda Mai, Hanadi El Achi, Jacob Armstrong, Amer Wahed, and Andy Nguyen**[*]

Department of Pathology and Laboratory Medicine, The University of Texas Health Science Center at Houston, Houston, TX, USA

[*]Corresponding Author: Andy Nguyen, M.D., Department of Pathology and Laboratory Medicine,

The University of Texas Health Science Center at Houston,

6431 Fannin, MSB 2.292, Houston, TX, 77030, USA,

Telephone: (713) 500–5337, Fax: (713) 500–0712; Email: nghia.d.nguyen@uth.tmc.edu

**ABSTRACT**

Recently, vision transformers (ViTs) were shown to be capable of outperforming convolutional neural networks when pretrained on sufficient amounts of data. In comparison to convolutional neural networks (CNNs), ViTs have a weaker inductive bias and therefore allow a more flexible feature detection. ViT models show good accuracy on large scale datasets, with features of self-supervised learning, and multi-modal training. Due to their promising feature detection, we explore ViTs for diagnosis of anaplastic large cell lymphoma (ALCL) vs. classical Hodgkin lymphoma (cHL) in digital pathology whole slide images (WSIs) of H&E slides. We compared the classification performance of the ViT model to our previously designed CNN on the same dataset. The dataset includes digital WSIs of H&E slides of 20 cases, including 10 cases for each diagnostic category (ALCL and cHL). From each WSI, 60 image patches (100x100 pixels) at 20x magnification were obtained to yield 1200 image patches, from which 1079 (90%) were used for training, 108 (9%) for validation, and 120 (10%) for testing. The test results from CNN model had previously shown an excellent diagnostic accuracy at 100% . The test results from the

ViT model showed a comparable accuracy at 100%. To the best of the authors' knowledge, this is the first direct comparison of predictive performance between a ViT model and a CNN model using the same dataset including cases of ALCL and cHL. The results show that the ViT's performance is comparable to that of the CNN model in diagnosis of ALCL vs. cHL using the same dataset of relatively small size (1200 images). Overall, CNNs are a more mature architecture than ViTs and are easier to work with due to existing frameworks and training recipes that are tried and tested. CNNs are the best choice when large scale pretraining is not an option. Nevertheless, our current study shows a comparable and excellent accuracy of ViT compared to that of CNN even with a relatively small dataset of ALCL and cHL.

**INTRODUCTION**

Machine learning and deep learning

Machine learning consists of software algorithms that can learn from and make predictions on data - i.e., "gives software the ability to learn without being explicitly programmed" [1]. Numerous machine learning methods have been attempted in the past with varying degree of success: Decision tree, Cluster analysis, Support vector machine, Random forest, Bayesian, Regression analysis, Neural network, large language model (LLM), etc. Deep learning is the most recent and most disruptive method of machine learning; based on Neural network and LLM [2,3]. Currently, many large companies are analyzing large volumes of data for business analysis and decisions, using Deep Learning technology (Google, Microsoft, openAI, etc.). The application of deep learning to digital pathology image has a promising start; it could impact personalized diagnostics, and treatment.

Major breakthroughs in Deep Learning started in 2006 and helped it to outperform all other machine learning models. Deep Learning algorithms [4] include two critical features: (1) Unsupervised learning allows a network to be fed with raw data (no known outcomes) and to automatically discover the representations needed for detection or classification, and (2) Extract high-level & complex data representations through multiple layers; avoid problems of last-gen networks. Deep learning has significantly benefited from supporting hardware [4] to support parallel computation, in the form of multiple graphics processing units (GPU).

Diagnosis of lymphoma using digital images

Lymphoma is a clonal malignancy of lymphocytes (either T- or B cells). The Classification of Lymphoid Malignancies (World Health Organization) includes at least 38 entities [5]. Lymphoid malignancies were diagnosed in 280,000 people annually worldwide. Lymphoma is typically first suspected by their pattern of growth and the cytologic features of the abnormal cells via light microscopy of hematoxylin-eosin stained tissue sections. Immunophenotypes are required for diagnosis (by flow cytometry and/or immunohistochemical stains). In addition, cytogenetics, molecular results, and clinical features are often needed in finalizing the diagnosis in certain lymphoma types.

Due to subtle difference in histologic findings between various types of lymphoma, histopathologic screen often presents a challenge to the pathologists. An automated diagnosis for digital images would be helpful to assist the pathologists in daily work. Previous attempts to digitally classify histologic images were based on specific criteria (such as nuclear shape, nuclear size, texture, etc.). They were not very successful [6]. Attention has turned to machine learning. In recent years, 'deep learning' techniques, especially convolutional neural network (CNN or ConvNet), has quickly become the state of the art in computer vision [7,8].

Convolution in CNN

Convolution is an operation in image processing using filters [9], to modify or detect certain characteristics of an image (Smooth, Sharpen, Intensify, Enhance). In CNN, it is used to extract features of images. Mathematically, a convolution is done by multiplying the pixels' value in image patch by a filter matrix (kernel matrix), to yield a dot product. By moving the filter across input image, one obtains the final output as a modified filtered image (Fig. 1).

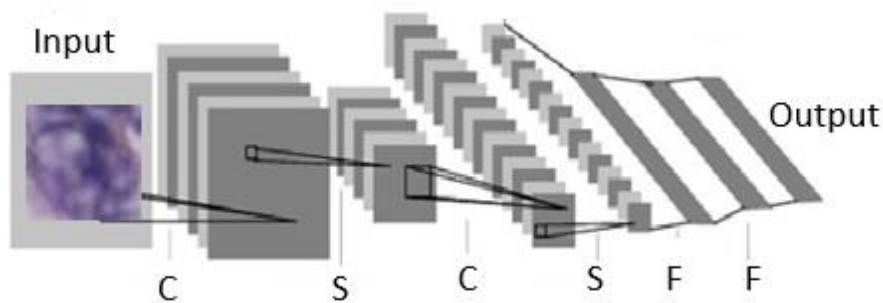

Fig. 1. Basic components of a Convolutional neural network: the convolutional layers (C)
perform 'feature extraction' consecutively from the image patch to higher level features;
the max pooling layers (S) reduce image size by subsampling; the last 'fully connected'
layers (F): provide prediction

## Vision Transformer Model

Recently, vision transformers (ViTs) were shown to be capable of outperforming convolutional neural networks when pretrained on sufficient amounts of data [10, 11, 12]. In comparison to convolutional neural networks (CNNs), vision transformers have a weaker locality bias and therefore allow a more general feature detection (multi-modal data). To track attention links between two input tokens, ViTs are used. The pixel is the most basic unit of measurement in an image, but calculating every pixel relationship in a normal image would be memory-intensive. ViTs, however, take several steps to do this, as described below (Fig. 2):

- -ViTs divide the full image into a grid of small image patches
- -ViTs apply linear projection to embed each patch, with consideration for position of each image patch in the image
- -Then, each embedded patch becomes a token, and the resulting sequence of embedded patches is passed to the transformer encoder
- -Then, the transformer encoder processes the input patches, using multi-head attention, and the output is fed into the multilayer perceptron (MLP), producing the resultant classes such as types of tumors in a tumor classifier.

Self-attention in ViT allows each part of image to relate (pay attention) to other parts of image, regardless of the distance between them. ViT has been used to build foundation models which are trained on enormous datasets using self-supervised learning, which do not require labeled labels [13]. They can be fine-tuned for a wide range of downstream tasks using a modest amount of task-specific labeled data for training.

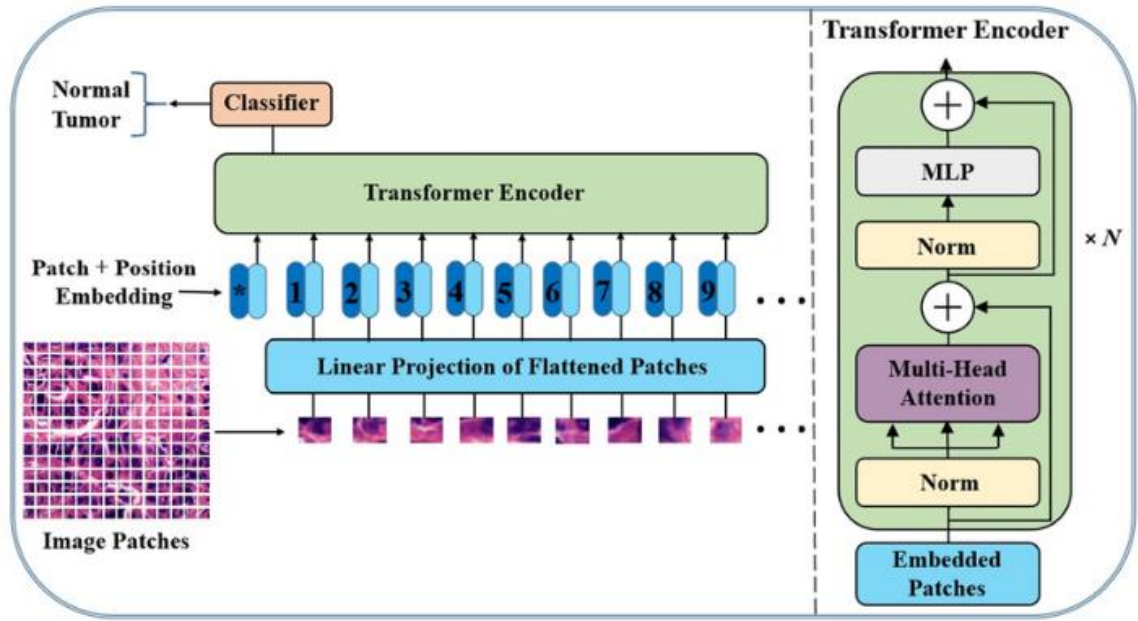

Fig. 2. The ViT model splits the image into a grid of non-overlapping patches before passing them to a linear projection layer as tokens. These tokens are then processed by a series of multi-headed self-attention layers to capture global relationship

Comparison of Accuracy by ViT to that by Convolutional Neural Network Model [10,11,12]

While the Transformer architecture (such as in ChatGPT, Copilot, etc., based on large language model) has become the de-facto standard for natural language processing tasks, its applications to computer vision remain limited. Then in 2021, Alexey Dosovitskiy et al presented the topic *"An image is worth 16x16 words: transformers for image recognition at scale"* at the International Conference on Learning Representations [12]. They applied ViT model directly to images and showed that the reliance on CNNs is not necessary and a pure transformer applied directly to sequences of image patches can perform very well on image classification tasks.

Due to their promising feature detection, this study explores ViTs for diagnosis of anaplastic large cell lymphoma (ALCL) vs. classical Hodgkin lymphoma (cHL) in pathology whole slide images (WSIs) of H&E slides. We will attempt to compare the classification performance of the ViT model to our previously designed CNN on the same dataset which has an excellent accuracy of 100% [14]. Note that a few previous studies showed that ViT models may be designed to yield predictive accuracy comparable to that of CNN with small datasets. These studies include: (a) renal pathology [15], by Zhang et al., (b) bladder pathology [16], by Ola S. Khedr et al., and (c) ImageNet-1k [17], by Lucas Beyer et al.

**MATERIALS AND METHODS**

We conducted a retrospective compilation of cases with newly diagnosed CHL and ALCL by current World Health Organization criteria at our institution from 2017 to 2024. We reviewed the morphological characteristics of each case and selected the hematoxylin and eosin- (H&E) stained slides from 20 cases, which were scanned using the SG60 scanner (Philips Corporation, Amsterdam, Netherlands) at 40x magnification (Fig. 3). The SG60 scanner has capacity for 60 glass slides, produces high-quality images, full automation (for focus, calibration, brightness and contrast settings), with tissue shape detection to outline and scan non-rectangular regions of interest for shorter turnaround times. The total scan time of a slide for a 15 × 15 mm benchmark scan area at a 40x resolution is ≤ 62 seconds. The images were acquired and stored in iSyntax2 format. Philips Image Management System was used to display the images. From each WSI, 60 image patches of 100x100 pixels (at 20x magnification, 0.5 μm/ pixel) were obtained for feature extraction with SnagIt software (TechSmith Corp, Okemos, Michigan, USA).

A total of 1200 image patches were obtained from which 1079 (90%) were used for training, 108 (9%) for validation, and 120 (10%) for testing. The cases were divided into two cohorts, with 10 cases for each diagnostic category.

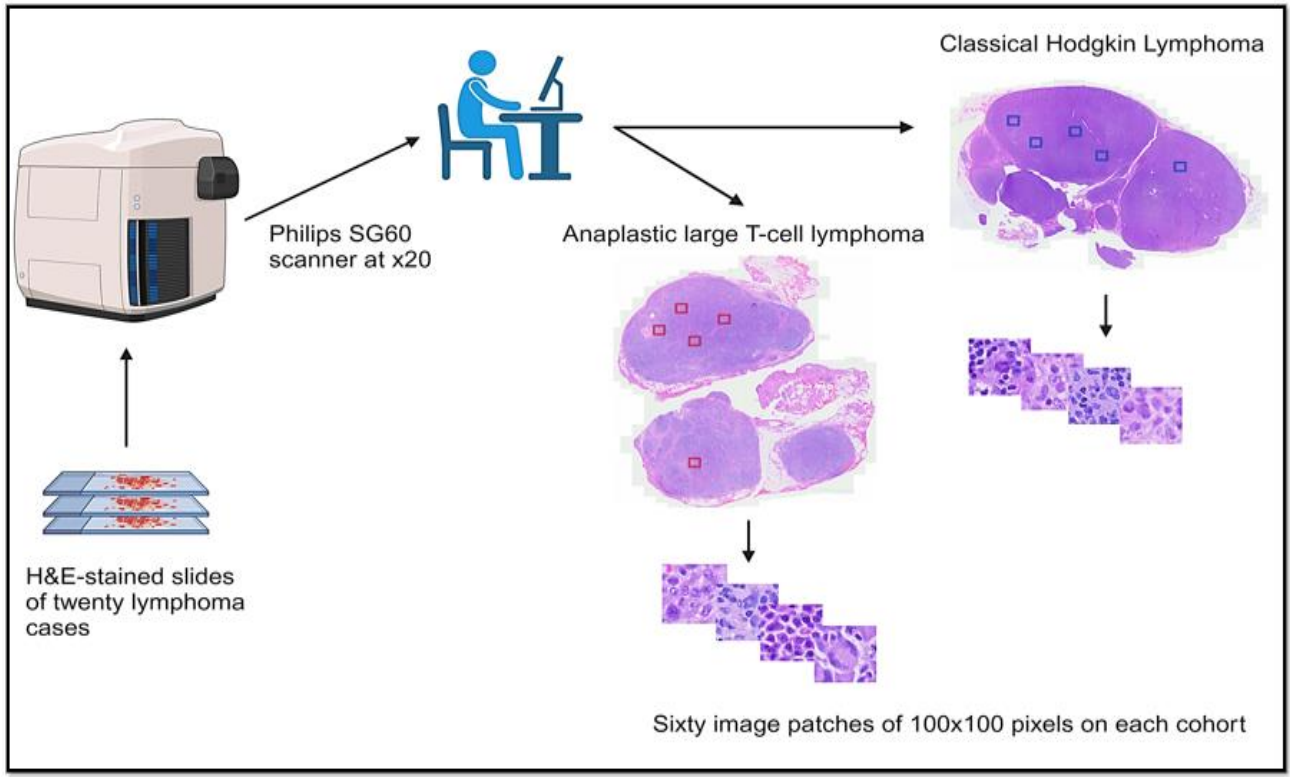

Fig. 3. Process of obtaining digital images for use in the study

Hardware platform for our model design:

CPU: Intel Xeon Gold 5222; 48 GB RAM (Intel Corp, Santa Clara, California, USA)

GPU: NVIDIA RTX A4000, 16 GB, 6144 CUDA cores (NVIDIA Corp, Santa Clara, California, USA)

Software platform for our model design:

Operating system: Windows 11 Professional, 64 bit (Microsoft Corp, Redmond, Washington, U.S.A.)

Language: Python, Torch, Torchvision (Python Software Foundation, Wilmington, Delaware, USA) to build ViT model (Table 1)

Hyperparameters used in the model:

- img_size = 100      # size of images, for example 100 for 100x100 images
- patch_size = 20      # size of image patches, for example 5 for 5x5 image patches
- d_model = 128      # the dimensionality of the model: commonly used values for d_model in practice are 128, 256, 512, or 1024
- num_heads = 4      # number of attention heads; it evenly divides the d_model dimension; i.e., d_model/num_heads=integer
- num_layers = 6      # number of transformer layers; commonly used values are between 6 and 12
- lr=0.001              # learning rate
- num_epochs = 200  # number of epochs
- num_classes = 2     # number of prediction classes
- batch_size=32       # number of cases in each reading batch for datasets
- dropout=0.1          # dropout rate for residual connections

Accuracy of prediction = Number of correct diagnoses/Number of all test images

(test images include 60 ALCL and 60 cHL)

Overfitting with small training dataset is minimized by:

- Minimizing Number of layers (6), image size (100), Number of attention heads (4)
- Use dropout layer in transformer block (0.1)

- Use large image patch size (20)
- Use optimizer Adam, which works better with transformers

Table 1. Components of the Visual Transformer Code

| PART 1 | ■ CORE CODE FOR THE ViT ENGINE:<br>    Initialize the model parameters<br>    Create a Vision Transformer model instance<br>    Positional Encoding with Sine and Cosine<br>    Multi-Head Attention<br>    Define query, key, and value<br>    Computes attention score (based on query and key), applies SoftMax on this score,<br>      and then computes the context vector (based on query, key, value)<br>    Concatenation results from multi-heads<br>    Transformer Block with ReLU (rectifier activation function),<br>      drop-out for residual connection<br>    Vision Transformer class |
|---|---|
| PART 2 | ■ Initialize Vision Transformer instance and Loss function/Optimizer (Adam) |
| PART 3 | ■ LOAD/PREPARE DATASET (TRAIN SET, TEST SET) |
| PART 4 | ■ TRAIN THE MODEL on the train set |
| PART 5 | ■ EVALUATE THE MODEL on the test set |
| PART 6 | ■ Display all test images with known diagnosis and predicted diagnosis |

**RESULTS**

The test results from our ViT model showed a diagnostic accuracy at 100% for 120 test images. Fig. 4 shows screen display with:

software introductory title, information on dataset, and loss function. Fig. 5 shows accuracy of prediction for 120 test images, along with display of each test image with its diagnosis and predicted diagnosis.

```
++++++++++++
VISION TRANSFORMER MODEL
Written in Python to assist diagnosing Anaplastic Large Cell Lymphoma versus
Classical Hodgkin Lymphoma

This model is based on pytorch, torchvision, and is designed with Multi-head Attention,
sine/cosine embedding function, and residual connections
.........
PARAMETERS OF MODEL:
Number of images in Trainning set:  1080
Number of images in Testing set:  120
Size of images:  100
Size of image patches:  20
Dimensionality of the model:  128
Number of attention heads:  4
Number of transformer layers:  6
learning rate:  0.001
Feedforward dimension:  2048
Number of epochs:  200
Number of prediction classes:  2
Number of cases in each reading batch for datasets:  32
++++++++++++
TRAINING THE VISON TRANSFORMER MODEL:
Loss in each epoch, with number of test images in each Batch:
[Epoch 1, Batch of: 30] loss: 0.234
[Epoch 2, Batch of: 30] loss: 0.203
[Epoch 3, Batch of: 30] loss: 0.201
[Epoch 4, Batch of: 30] loss: 0.201
[Epoch 5, Batch of: 30] loss: 0.201
[Epoch 6, Batch of: 30] loss: 0.197
[Epoch 7, Batch of: 30] loss: 0.198
[Epoch 8, Batch of: 30] loss: 0.200
[Epoch 9, Batch of: 30] loss: 0.197
[Epoch 10, Batch of: 30] loss: 0.187
[Epoch 11, Batch of: 30] loss: 0.187
[Epoch 12, Batch of: 30] loss: 0.178
[Epoch 13, Batch of: 30] loss: 0.197
[Epoch 14, Batch of: 30] loss: 0.177
[Epoch 15, Batch of: 30] loss: 0.171
[Epoch 16, Batch of: 30] loss: 0.177
[Epoch 17, Batch of: 30] loss: 0.168
[Epoch 18, Batch of: 30] loss: 0.165
[Epoch 19, Batch of: 30] loss: 0.170
[Epoch 20, Batch of: 30] loss: 0.158
```

← Display of software introductory info

← Display of dataset info, hyperparameters used in execution

← Display of Loss in each Epoch

Fig. 4. Screen display of model parameters

Fig. 5. Screen display of prediction results

A production protocol for testing unknown images is established for the ViT trained model to offer prediction of diagnosis for new (unknown) images.
Fig. 6 illustrates a typical screen display for 2 unknown images (Test01.jpg and Test02.jpg). They were both correctly predicted to be ALCL and cHL, respectively.

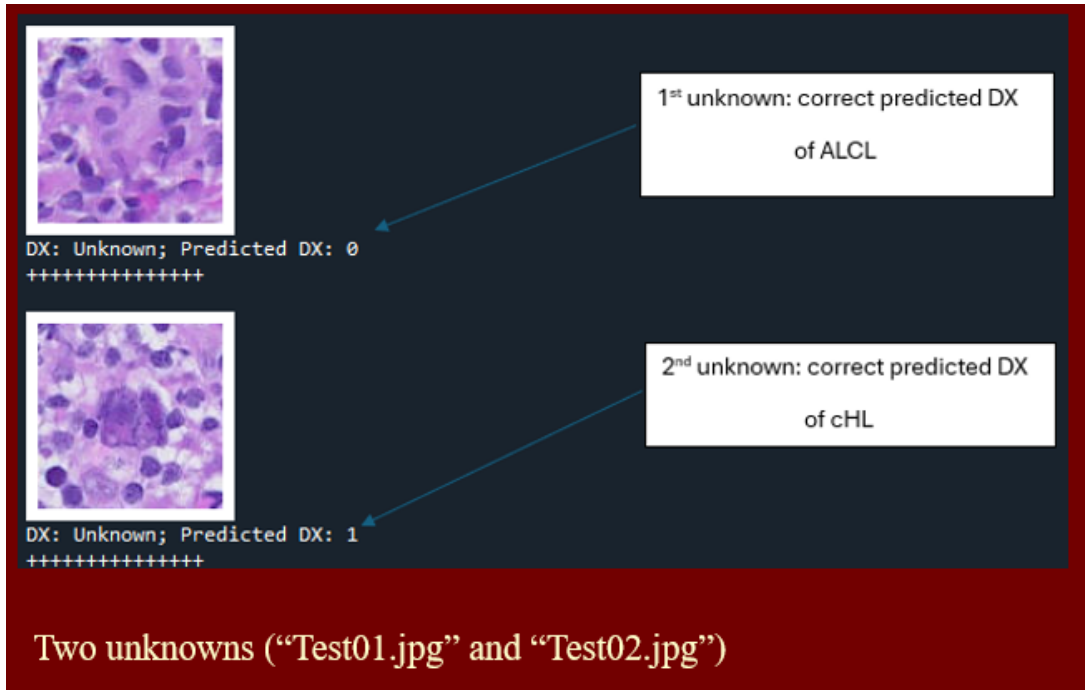

Fig. 6. Screen display of the production model, predicting diagnosis for 2 unknown images

**DISCUSSION**

To the best of the authors' knowledge, this is the first direct comparison of predictive performance between a ViT model and a CNN model using the same dataset including cases of ALCL and cHL cases. The re sults show that the ViT model's performance is the same as that of the CNN model (100%) in diagnosis of ALCL vs. cHL using the same (and relatively small) dataset of 1200 images. Possible reasons for the high accuracy of ViT in this study are likely due to: (a) highly accurate labelling of image patches by hematopathologists, (b) rigid inclusion constraints of WSI (by the same scanner in the same lab), (c) only 2 types of lymphomas to classify the test images into, and (d) cytology of malignant cells does not vary over large distance in sections of WSI.

From literature review [10-13], CNNs are a more mature architecture than ViTs, which can make it easier to work with due to existing frameworks and training recipes that have been tried and tested. CNNs are usually the best choice when large scale pretraining is not an option. CNNs are a practical and high-performing choice for many real-world applications. ViTs are thought to benefit from the lack of strong inductive biases exhibited by traditional CNN models, including inherently spatial, local and hierarchical feature processing operations. In the ViT models, the absence of many of these convolution-like inductive biases can lead to improved generalizability. On the other hand, the lack of inductive bias presents unique challenges, as such models typically require very large quantities of data to train on. Transformer-based architectures are computationally expensive due to the computation of the self-attention mechanism, which is usually quadratic to the size of the input image. This issue appears to be less of a problem with natural images, but with histopathological images, it is a significant difficulty since WSIs come in gigapixels and are larger in size compared to natural image datasets. However, our current study shows comparable performance between CNN model and ViT model even with smaller datasets such as the one we use here for ALCL and cHL.

Limitation of current study includes using only labeled data with supervised training for ViT. Plan for unsupervised training (such as in foundation models) will be considered in future studies. Future studies will also compare CNN and ViT models on much larger datasets and check for consistent accuracy of ViT model on various stains and scanner platforms.

## SUMMARY

In the ViT models, the absence of locality biases can lead to improved generalizability. On the other hand, they typically require very large quantities of data to train on. Transformer-based architectures are computationally expensive due to the computation of the self-attention mechanism, especially with the size of histopathological images, such as WSIs in gigapixels. Despite requirement of large training dataset by ViT models as reported in literature, our study showed that ViT models can be designed to yield the same prediction accuracy as that of CNN models for our relatively small dataset of ALCL and cHL. In summary, ViT models are worth keeping an eye on due to reported good accuracy on large scale datasets, self-supervised learning, and multi-modal tasks. We anticipate further development in reducing transformer computational complexity in the near future.

**ACKNOWLEDGEMENT** (Projects on Deep Learning in Pathology)
- Faculty at U of Texas-Houston, Medical Scholl, Pathology:

    L Chen, MD

    Z Hu, MD

    H El Achi, MD

    A Wahed, MD

    I Wang, MD

    B Mai, MD

    J Armstrong, MD
- Faculty at U of Texas-Houston, Medical Scholl, Oncology:

    A Rios, MD

    Z Kanaan, MD

- Residents at U of Texas-Houston, Medical Scholl, Pathology:
    T Belousova, MD
    D Rivera, MD
    A Banerjee, DO
- Staff in Digital Pathology Laboratory, U of Texas-Houston, Medical Scholl, Pathology:
    K Ali, BS
    R Zhang, MD
    Jacob Huddin, MD

**REFERENCES**

1. Razzak M.I., Naz S., Zaib A. Deep Learning for Medical Image Processing: Overview, Challenges and the Future. In: Dey N., Ashour A., Borra S. (eds) Classification in BioApps. First ed. Springer International Publishing; 2018:323-350

2. MIT Technol. Rev., 2013. Available at (last accessed on 10/30/18):
https://www.technologyreview.com/s/513696/ deep-learning

3. LeCun, Y., Bengio, Y. & Hinton, G. Deep learning. Nature. 2015;521:436–444

4. Andrew Janowczyk, Anant Madabhushi. Deep learning for digital pathology image analysis: A comprehensive tutorial with selected use cases. J Pathol Inform. 2016;7:29.

5. WHO Classification of Tumours: Haematolymphoid Tumours, 5th Edition, Volume 11, 2024. WHO Classification of Tumours Editorial Board. 69008 Lyon, France: International Agency for Research on Cancer (IARC)

6. Choras RS. Feature extraction for CBIR and biometrics applications. 7th WSEAS International Conference on Applied Computer Science. Vol. 7. 2007

7. Marsland S. Machine learning: an algorithmic perspective. Chapman and Hall/CRC, 2011.

8. Mitchell TM, Mitchell TM. Machine learning. Vol. 1. No. 9. New York: McGraw-Hill, 1997

9. Roy K, Banik D, Bhattacharjee D, Nasipuri M. Patch-based system for classification of breast histology images using deep learning. Computerized Medical Imaging and Graphics 71 (2019): 90-103.

10. Asim Waqas, Marilyn M. Bui, Eric F. Glassy et al. Revolutionizing Digital Pathology With the Power of Generative Artificial Intelligence and Foundation Models. Laboratory Investigation. Volume 103, Issue 11,100255, November 2023

11. Ashish Vaswani, Noam Shazeer, Niki Parmar et al. Attention Is All You Need . 31st Conference on Neural Information Processing Systems (NIPS 2017)

12. Alexey Dosovitskiy , Lucas Beyer , Alexander Kolesnikov et al. AN IMAGE IS WORTH 16X16 WORDS: TRANSFORMERS FOR IMAGE RECOGNITION AT SCALE. Proceeding of ICLR 2021

13. Eugene Vorontsov, Alican Bozkurt, Adam Casson, George Shaikovski, Michal Zelechowski, Kristen Severson et al. A foundation model for clinical-grade computational pathology and rare cancers detection. Nature Medicine | Volume 30 | October 2024 | 2924–2935

14. Daniel Rivera, Kristine Ali, Rongzhen Zhang, Brenda Mai, Hanadi El Achi, Jacob Armstrong, Amer Wahed, and Andy Nguyen. Deep Learning-Based Morphological Classification between Classical Hodgkin Lymphoma and Anaplastic Large Cell Lymphoma: A Proof of Concept and Literature Review, 21st Century Pathology, Volume 4 (1): 159

15. Ji Zhang, Jia Dan Lu, Bo Chen et al. Vision transformer introduces a new vitality to the classification of renal pathology. BMC Nephrology (2024) 25:337

16. Ola S. Khedr, Mohamed E. Wahed, Al-Sayed R. Al-Attar et al. The classification of the bladder cancer based on Vision Transformers (ViT). Nature Scientific Reports | (2023) 13:20639

17. Lucas Beyer et al. Better plain ViT baselines for ImageNet-1k. Google Brain Research, Zurich https://github.com/google-research/big_vision